\documentclass[journal]{IEEEtran}
\usepackage{url}
\usepackage{cite}
\usepackage{amsmath,graphicx,siunitx}
\usepackage{float}
\usepackage{multirow}
\usepackage{threeparttable}
\usepackage{booktabs}
\usepackage{makecell}
\usepackage[justification=centering]{subcaption}
\usepackage{amssymb}
\usepackage{color,xcolor,soul}
\usepackage{caption} 
\usepackage{flushend}
\usepackage{hhline}
\usepackage{ulem}
\usepackage{gensymb}

\ifCLASSINFOpdf
\else
\fi
\begin{document}
%
\title{SIGAN: A Novel Image Generation Method for Solar Cell Defect Segmentation and Augmentation}
%
%
%

\author{Binyi Su, Zhong Zhou, Haiyong Chen, and Xiaochun Cao, \textit{Senior Member, IEEE} 
	\vspace{-2em}
	\thanks{This work was supported by National Key R$\&$D Program of China under Grant 2018YFB2100601, Natural Science Foundation of China under Grant 61872023 and Grant 62073117, and Central Government Guides Local Science and Technology Development under Grant 206Z1701G. (\textit{Corresponding author: Zhong Zhou.})
		
	B. Su, Z. Zhou are with the State Key Laboratory of Virtual Reality Technology and Systems, School of Computer Science and Engineering, Beihang University, Beijing 100191, China (e-mail: Subinyi@buaa.edu.cn; zz@buaa.edu.cn).
		
	H. Chen is with the School of Artificial Intelligence and Data Science, Hebei University of Technology, Tianjin 300130, China (e-mail: haiyong.chen@hebut.edu.cn).	
		
	Xiaochun Cao is with the State Key Laboratory of Information Security, Institute of Information Engineering, Chinese Academy of Sciences, Beijing 100093, China, and also with Peng Cheng Laboratory, Shenzhen 518055, China (e-mail: caoxiaochun@iie.ac.cn).}}

%
%

\markboth{Journal of \LaTeX\ Class Files,~Vol.~14, No.~8, August~2015}%
{Shell \MakeLowercase{\textit{et al.}}: Bare Demo of IEEEtran.cls for IEEE Journals}
%



\maketitle

\begin{abstract}
Solar cell electroluminescence (EL) defect segmentation is an interesting and challenging topic. Many methods have been proposed for EL defect detection, but these methods are still unsatisfactory due to the diversity of the defect and background. In this paper, we provide a new idea of using generative adversarial network (GAN) for defect segmentation. Firstly, the GAN-based method removes the defect region in the input defective image to get a defect-free image, while keeping the background almost unchanged. Then, the subtracted image is obtained by making difference between the defective input image with the generated defect-free image. Finally, the defect region can be segmented through thresholding the subtracted image. To keep the background unchanged before and after image generation, we propose a novel strong identity GAN (SIGAN), which adopts a novel strong identity loss to constraint the background consistency. The SIGAN can be used not only for defect segmentation, but also small-samples defective dataset augmentation. Moreover, we release a new solar cell EL image dataset named as EL-2019, which includes three types of images: crack, finger interruption and defect-free. Experiments on EL-2019 dataset show that the proposed method achieves 90.34\% F-score, which outperforms many state-of-the-art methods in terms of solar cell defects segmentation results.


\end{abstract}


%
\IEEEpeerreviewmaketitle

\section{Introduction}
%
%
%
%
\IEEEPARstart{T}{oday}, when non-renewable energy such as oil and coal are about to be exhausted, solar energy is becoming the focus of the world. Photovoltaic solar cells are the main products that convert solar energy into electric energy. In the process of intelligent manufacturing, the solar cell defect inspection is an essential part that can guarantee the products with high quality \cite{2020Hu,Su2020,Su2021}. Automated solar cell defect detection is one of the most direct applications of artificial intelligence algorithms in an industrial setting. Moreover, solar cell defect detection is always of considerable interest
to both industrial and academic researchers. Traditional methods \cite{Su2019,Yan2020} mainly rely on filters or feature descriptors, which have some apparent limitations. One is the poor generalization performance, we need to select a specific filter or feature descriptor for a specific task, and some prior knowledge is necessary during this process. Another is weak ability of anti-disturbance, especially in the complex situation of industrial applications. These approaches are easily disturbed by illumination, imaging quality, and complex background. Thus, traditional methods are inevitably replaced by new and better algorithms.

\begin{figure}[!t]
	\centering
	\includegraphics[width=5.5cm]{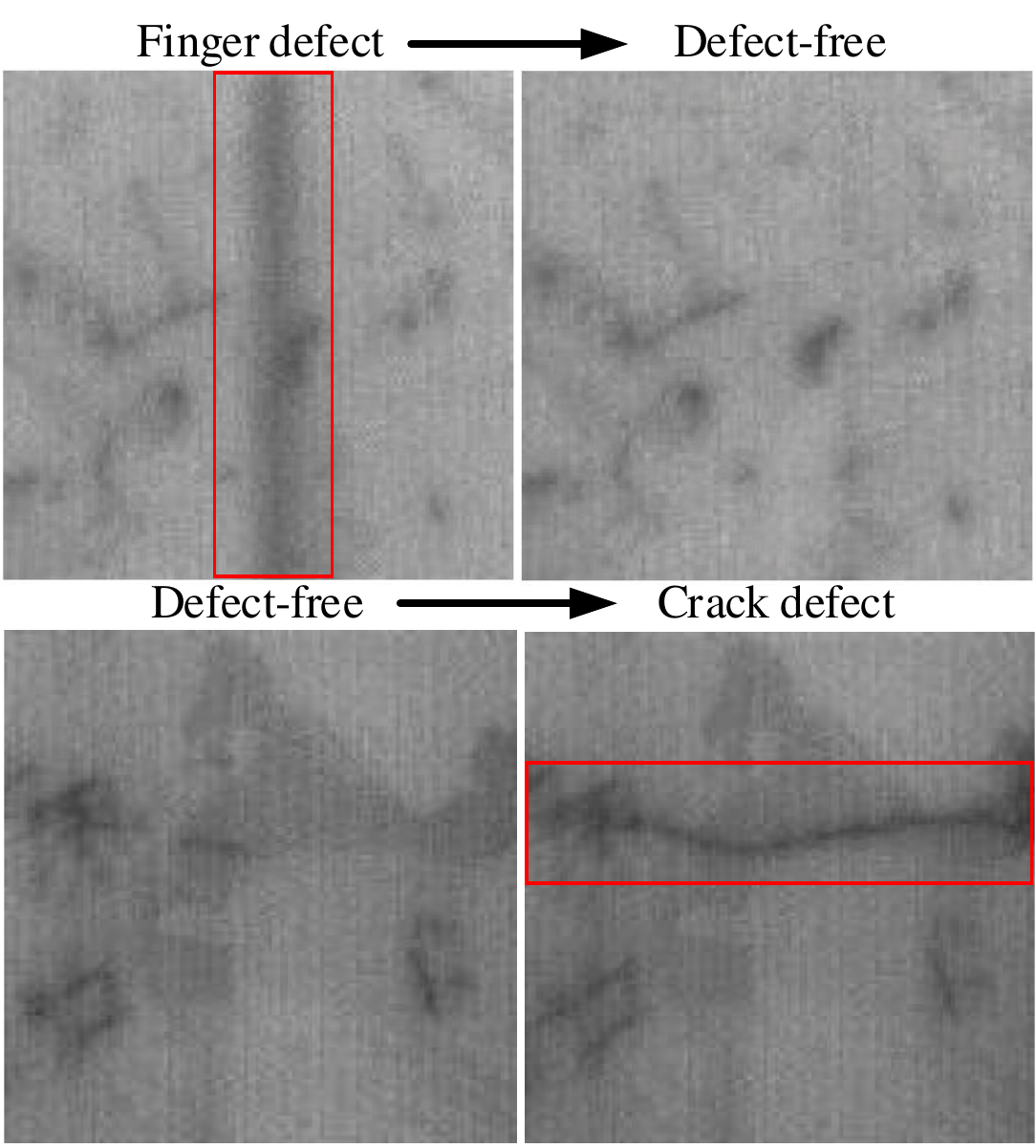}
	\caption{Mutual transformation between defective and defect-free images, while keeping the background almost unchanged.}\label{fig1}
\end{figure}

Recently, deep learning has arose many scholars attention. The advantages of deep learning present as high accuracy, good generalization performance, and strong anti-interference ability. It has obtained many successful industrial applications, such as face recognition \cite{He2020}, cityscape segmentation \cite{Geng2021}, style transfer \cite{Cheng2020} and defect detection \cite{Su2021,BAF_detector2021}. Deep learning algorithm will automatically extract the semantic and texture feature of the input image, without designing a specific feature extractor for different tasks. 

The deep learning model based on generative adversarial network (GAN) \cite{Lan2014,Li2019,Zhang2020a} to eliminate image noise is excellent, which can filter out the noise disturbance and retain image details well. Another successful application of GAN is image generation \cite{Wang2018a,Rout2020,Yi2020}, which can be roughly divided into the following two parts: object to object (style transfer) and object to object-free (or object-free to object). The first one is usually called style transfer \cite{Zhang2020}, which migrates some attributes of the object in the source domain to the object in the target domain, such as apple to orange, zebra to horse, summer to winter. The second one is from ``yes" to ``no" (or from ``no" to ``yes") \cite{Liu2020, Niu2020}, which means that an input image contains the object, which will be removed in the output image of the GAN. An example of the proposed algorithm is shown in the first row of Fig. \ref{fig1}.

Yes2no (defect2defect-free) can be viewed as filtering, after inputting a defective image to the GAN, the generator of GAN will filter out the defect and almost completely preserve the background texture simultaneously. Then, we can make difference between the generated image with the input image to eliminate the background. Only the defect region is retained. By thresholding the eliminated image, the defect region can be segmented. 

No2yes (defect-free2defect) can be viewed as defective dataset augmentation, which plays an important role in the industrial defect inspection task. Industrial datasets are different from natural scene datasets such as ImageNet dataset \cite{Jia2009} and COCO dataset \cite{Vinyals2016}. In the industrial manufacturing process, defect samples are difficult to collect, which account for a small part. However, defect-free samples are easy to collect, thus utilizing defect-free samples to generate realistic defect samples is very valuable for small-samples defect detection.

To accomplish above two tasks by an algorithm, some problems should be considered. Firstly, it is difficult to obtain one-to-one defective image and defect-free image annotation, so the supervised learning network can not be trained. This problem can be solved by the Cycle Consistency Generative Adversarial Network (CycleGAN) \cite{Zhu2017}, which can employ adversarial loss and cycle-consistency loss to carry out the mutual transformation of two image domains (defect and defect-free) without requiring pixel-wise annotation. Secondly, another key problem of the yes2no or no2yes is how to keep the background unchanged, whether it is to remove or generate defects. To overcome this problem, we propose the strong identity (SI) loss to constrain the textural similarity between input image and the generated image. In other words, except for the defect region, the background texture is as consistent as possible before and after image generation, as shown in Fig. \ref{fig1}. The main contributions of this paper can be summarized as follows: 

\begin{enumerate}
	\item A new image generation algorithm named as Strong Identity GAN (SIGAN) is proposed by employing a new strong identity (SI) loss to generate high-quality defective or defect-free image, while keeping the background almost unchanged. 
	\item This paper provides a new idea of using GAN for defect segmentation, which is accomplished by thresholding the difference between the input defective image and the generated defect-free image.
	\item SIGAN can be used not only for defect segmentation, but also small-samples defective dataset augmentation. Experimental results show that the defect segmentation method achieves the better performance than many state-of-the-art methods. Moreover, defect classification models trained with dataset augmented by the SIGAN perform substantially better than those trained without augmentation.  
	\item Almost all public industrial image datasets only contain defective images, they ignore the value of defect-free images. This paper releases a solar cell EL image dataset named as EL-2019, which contains 260 defective images and 280 defect-free images.
\end{enumerate}

This paper is organized as follows:  Section \uppercase\expandafter{\romannumeral2} presents an overview of the related works. Section \uppercase\expandafter{\romannumeral3} gives the details of the proposed methods. Section \uppercase\expandafter{\romannumeral4} presents extensive experiments. Finally, Section \uppercase\expandafter{\romannumeral5} concludes this study.

\section{Related Works}
\subsection{Defect Segmentation}
Defect segmentation methods can be roughly divided into two categories: one is the filter-based methods, the other is the CNN-based methods. 

In terms of filter-based approaches, the filter is employed to filter out the high frequency information or low frequency information. After filtering out the defect, the filtered image will make difference with the original input image to accomplish defect segmentation. Of course, the filter can also directly filter out the background and retain the defects simultaneously. Based on above strategies, some filter-based approaches have been proposed to segment the defect in EL images. Tsai \textit{et al.} \cite{Tsai2012} introduced the fourier image reconstruction technique to detect defect of solar cell EL image. The spatial image is firstly transformed into spectral image. Next, the spectral frequency associated with the defect was removed by setting the threshold. Finally, the defect region can be easily identified by evaluating the gray-level differences between the original image and its reconstructed image. Anwer \textit{et al.} \cite{Anwar2014} proposed an improved anisotropic diffusion filter, which can smooth the image while preserving the edge texture. This method performs well in micro-crack defect detection task. Chen \textit{et al.} \cite{Chen2019} applied 2D Hessian-based enhancement filter to obtain the linear-structure and blob-structure. Then, a novel structure similarity measure (SSM) function is designed by using the identification functions of two structures, which can highlight crack defect (line-structure), and suppress crystal grains (blob-structure) simultaneously. This approach is relatively effective in micro-crack defect identification and outperforms the previous methods. Subsequently, Chen \textit{et al.} \cite{Chen2020} proposed a novel steerable evidence filter (SEF) to detect crack defects in solar cell EL images. SEF is an oriented filter, which is robust to the arbitrary texture orientations. Thus, it is better to detect various shapes of crack defects. 

Recently, Convolutional Neural Network (CNN) has achieved good performance in image segmentation task, researchers proposed many excellent algorithms \cite{Wang2019,2019Yu,2019Nakazawa,Ronneberger2015,Chen2018,Fu2019}, such as  UNet \cite{Ronneberger2015}, deeplabv3 \cite{Chen2018}, and DANet \cite{Fu2019}. Different from above supervised methods that requires pixel-wise annotation dataset for training, the approach proposed by this paper is weakly supervised and based on image generation. SIGAN does not need one-to-one corresponding data annotation, only needs to provide two image domains (defect and defect-free) for training. It is similar as the filter-based method. To accomplish the defect segmentation, SIGAN will employ deep learning network to filter out the defect region and retain the background simultaneously. It uses GAN to simulate the filters and complete the defect segmentation.


\begin{figure}[!t]
	\centering
	\includegraphics[width=8.7cm]{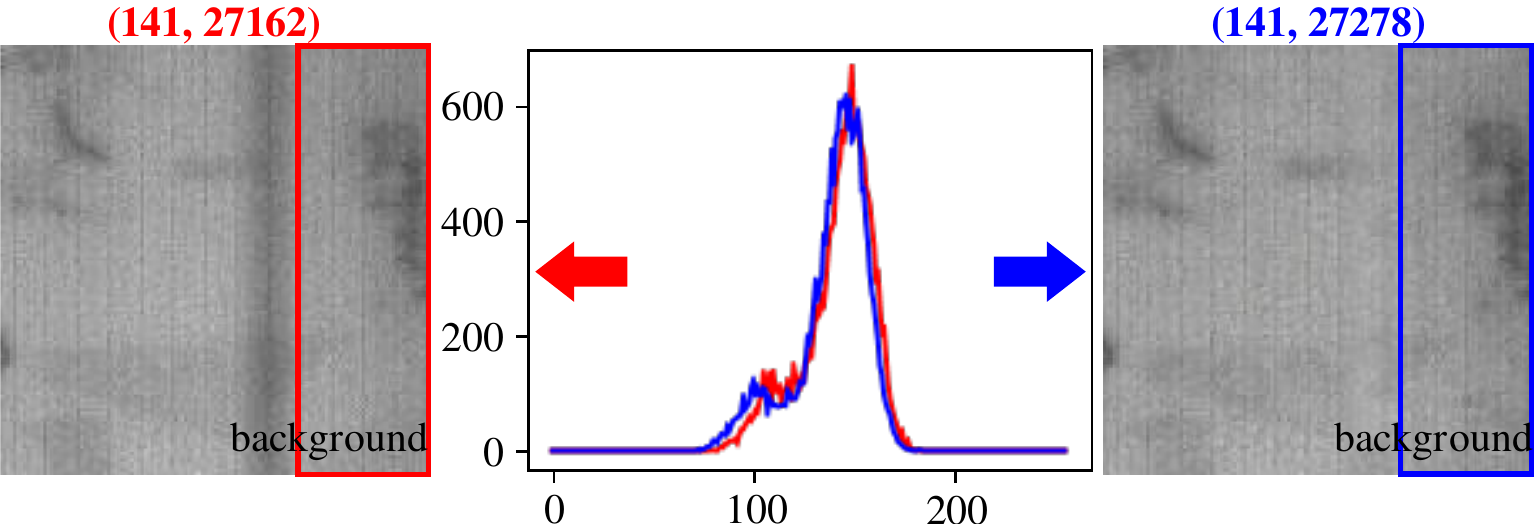}
	\caption{Gray histogram, brightness and contrast of the same background region for the finger interruption defect image and the generated defect-free image.}\label{fig3}
\end{figure}

\subsection{Image Generation}
Many researches have employed GAN-based image generation approaches \cite{Kun2020, Yu2019,Liu2020, Niu2020} to augment the image dataset. Liu \textit{et al.} \cite{Liu2020} proposed a multi-stage GAN to generate defective fabric images. They applied conditional GAN to synthesize the reasonable defective patches. Next, another GAN-based model is employed to fuse these patches into a raw large-resolution defect-free image. The generated defective images are expanded to augment the dataset, which can fine-tuning the segmentation network to better segment the defects. To improve the surface defect recognition, Niu \textit{et al.} \cite{Niu2020} proposed a surface defect-generation adversarial network (SDGAN) to augment the defect dataset. SDGAN employed the defect-free image to generate high-quality and diversity defective images, which were used for fine-tuning the classification effect of the surface defect.

Different from above methods, this paper provides a new idea of using GAN for solar cell defect segmentation, which is accomplished by thresholding the difference between the input defective image and the generated defect-free image. However, a key problem is that how to ensure the consistency of the background except for the defect area before and after the image generation. To solve this problem, this paper proposes strong identity (SI) loss in GAN to constrain the background details of generated image to be as similar as the input image except for the defect region. As shown in Fig. \ref{fig3}, the gray histogram, brightness and contrast of the same background region are close to each other before and after image generation with the proposed SIGAN. Thus, it is not hard to see that in the process of removing the defective area, the background remains almost unchanged, which verifies its effectiveness. Of course, the proposed SIGAN can accomplish the mutual transformation between defective and defect-free images, it can also be used for defective dataset augmentation (Inputting a defect-free image, SIGAN can generate a defective image.), which is validated by image classification results in this paper.


\section{Methodology}

This section firstly introduces the architecture of the proposed SIGAN, which can generate the defective or defect-free images. Next, we present the details of the defect segmentation approach based on defect2defect-free and defect augmentation approach based on defect-free2defect.


\subsection{The SIGAN}
The goal of the proposed SIGAN is to accomplish the mutual transformation between the defective image and defect-free image (defect2defect-free and defect-free2defect), while keeping the image background almost unchanged. SIGAN includes three types of losses: adversarial loss \cite{Lan2014} is used for matching the distribution of generated image to the data distribution in the target domain; Cycle consistency loss \cite{Zhu2017} guarantees the mutual transformation between defective image and defect-free image; The proposed strong identity (SI) loss constraints the generated image to be as similar as the input image except for the defect region. Next, the network architectures of the generator and discriminator are introduced in details.

\begin{figure*}[!t]
	\centering
	\includegraphics[width=15cm]{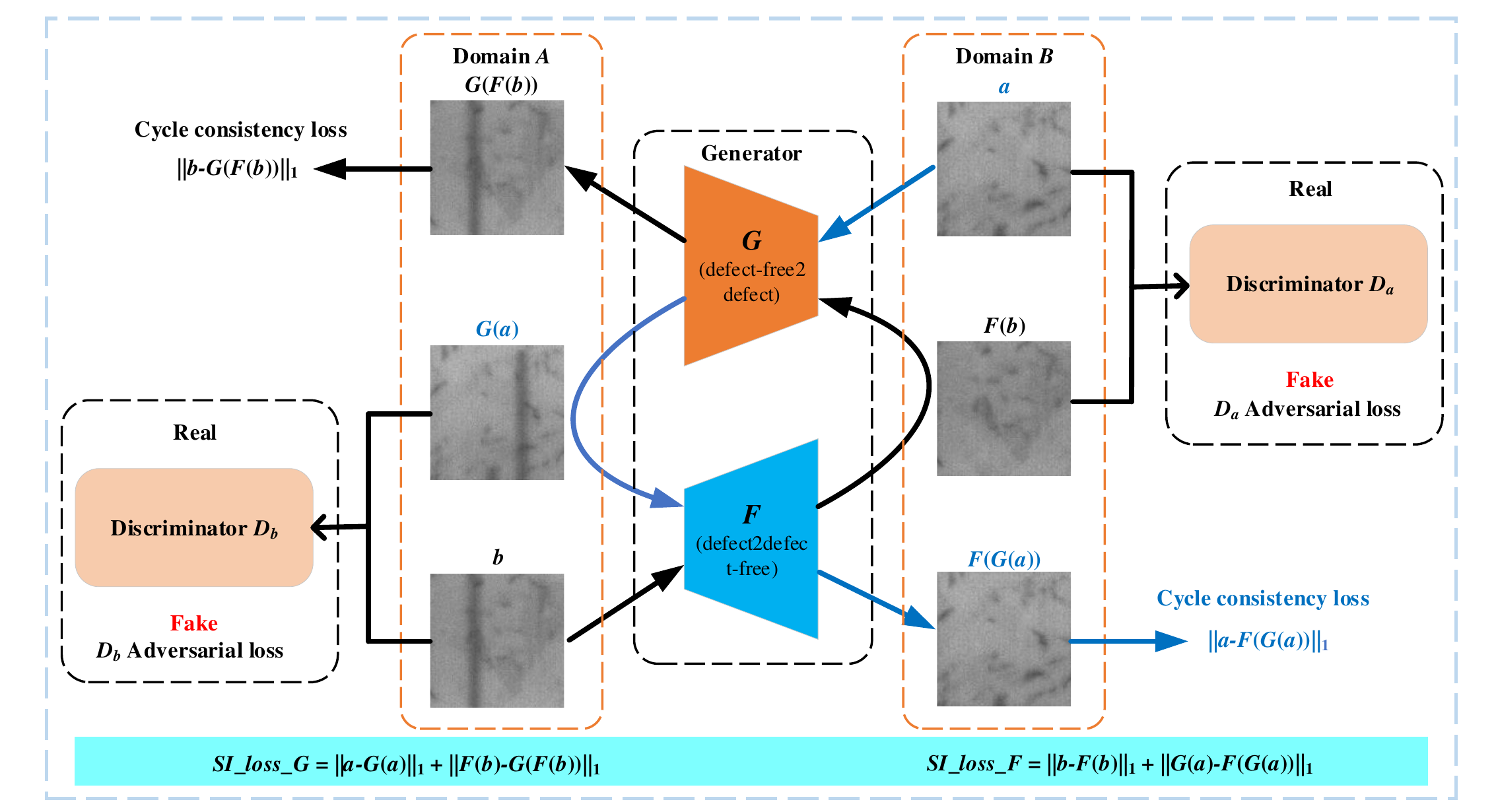}
	\caption{The architecture of the SIGAN.}\label{fig5}
\end{figure*}

\subsubsection{Adversarial Loss}
Adversarial loss \cite{Lan2014} plays a vital role to the cross domain image generation. For the source domain \textit{A} to target domain \textit{B}, the adversarial loss is defined as follows:

\begin{equation}\label{eq1}
	\begin{split}
		\begin{array}{l}L_{adv}(G,\;D_b,\;A,\;B)={\mathbb{R}}_{b\in P_{r}(b)}\lbrack\log D_b(b)\rbrack\\\;\;\;\;\;\;\;\;\;\;\;\;\;\;\;\;\;\;\;\;\;\;\;\;\;\;\;\;\;\;\;\;+\;{\mathbb{R}}_{a\in P_{r}(a)}\lbrack\log(1-D_b(G(a)))\rbrack\end{array}
	\end{split}
\end{equation}
where $P_{r}(a)$ denotes the image distribution of domain A,  $P_{r}(b)$ denotes the image distribution of domain B. As shown in Fig. \ref{fig5}, given two image domains: source domain \textit{A} and target domain \textit{B}, generator G tries to make the generated image $G(a)$ of source domain \textit{A} to be as similar as the image in target domain \textit{B}. While discriminator $D_b$ aims to distinguish the generated image $G(a)$ with real domain image $b$. In short, the generator G expects to cheat the discriminator $D_b$, but the discriminator $D_b$ does not allow the generated fake image $G(a)$ to be discriminated as b. This is a min-max problem, i.e., $\min_G\max_{D_b}L_{adv}(G,\;D_b,\;A,\;B)$. The loss optimization direction is to minimize $G$ and maximize $D_b$ in the training process of the novel SIGAN. Moreover, the mapping process from source domain \textit{B} to target domain \textit{A} is similar as above, i.e., $\min_F\max_{D_a}L_{adv}(F,\;D_a,\;B,\;A)$.  

\subsubsection{Cycle consistency Loss}
Cycle consistency loss \cite{Zhu2017} can ensure the SIGAN to accomplish the mutual transformation between domain \textit{A} and \textit{B} simultaneously: 1) defect-free domain \textit{A} to defective domain \textit{B}, 2) defective domain \textit{B} to defect-free domain \textit{A}. It is defined as follows:

\begin{equation}\label{eq2}
	\begin{split}
		\begin{array}{l}L_{cyc}(G,\;F)={\mathbb{R}}_{(a\in P_r(a))}{\left|\left|F(G(a))-a\right|\right|}_1\\\;\;\;\;\;\;\;\;\;\;\;\;\;\;\;\;\;\;\;+\;{\mathbb{R}}_{(b\in P_r(b))}{\left|\left|G(F(b))-b\right|\right|}_1\end{array}
	\end{split}
\end{equation}
where $\vert\vert\cdot\vert\vert_1$ denotes L1 norm.
For example, $a\rightarrow G(a)\rightarrow F(G(a))\approx a$ is a cycle, which can bring the generated image $G(a)$ by generator $G$ back to original image through generator $F$, as shown in Fig. \ref{fig5}. We hope that the generated image $F(G(a))$ should be as similar as the original input image $a$. Thus, L1 norm is employed to measure cycle consistency between $a$ and $F(G(a))$. Moreover, the reverse cycle $b\rightarrow F(b)\rightarrow G(F(b))\approx b$ should also be satisfied. 

For an image, the ground truth is not required in the training process of SIGAN, instead, two domains (defect and defect-free) are needed. Cycle consistency loss is one of the key point to ensure that two different domains can be transformed without one-to-one corresponding data annotation for training.

\subsubsection{Strong Identity Loss} 
No matter to generate defective image or defect-free image, a key problem for solar cell EL image generation is that how to ensure the consistency of the background except for the defect area before and after the image generation. To solve this problem, this paper proposes strong identity (SI) loss to limit the background of generated image to be as similar as the input image. SI loss plays a vital role in the image segmentation task, which applies the original defective image to subtract the generated defect-free image, then the defect region is retained and the background region is eliminated. For generator $G$, the SI loss is denoted as:

\begin{equation}\label{eq3}
	\begin{split}
		\begin{array}{l}L_{SI}(G,\;A,\;B)={\mathbb{R}}_{(a\in P_r(a))}{\left|\left|a-G(a)\right|\right|}_1\\\;\;\;\;\;\;\;\;\;\;\;\;\;\;\;\;\;\;\;\;\;\;\;\;+\;{\mathbb{R}}_{(b\in P_r(b))}{\left|\left|F(b)-G(F(b))\right|\right|}_1\end{array}
	\end{split}
\end{equation}
where $a$ and $F(b)$ are the inputs, $G(a)$ and $G(F(b))$ are the outputs of the generator $G$. For generator $F$, the SI loss is denoted as:

\begin{equation}\label{eq4}
	\begin{split}
		\begin{array}{l}L_{SI}(F,\;A,\;B)={\mathbb{R}}_{(b\in P_r(b))}{\left|\left|b-F(b)\right|\right|}_1\\\;\;\;\;\;\;\;\;\;\;\;\;\;\;\;\;\;\;\;\;\;\;\;+\;{\mathbb{R}}_{(a\in P_r(a))}{\left|\left|G(a)-F(G(a))\right|\right|}_1\end{array}
	\end{split}
\end{equation}
where $b$ and $G(a)$ are the inputs, $F(b)$ and $F(G(a))$ are the outputs of the generator $F$. In the experiments, the L1 norm is utilized to measure the SI loss. There are two reasons why we use the proposed SI loss to measure the input image and the output image of each generator to keep the background almost unchanged. Firstly, the adversarial loss magnifies the difference between the generated image and the original input image to accomplish the domain transformation. However, the SI loss is employed to make the difference smaller, and finally achieves a compromise balance with the adversarial loss.
Secondly, as shown in Fig. \ref{fig3}, the gray histogram of the input image is similar as the generated defective image of SIGAN. In terms of each generator, we hope that the data distribution of input image and output image should be as consistent as possible. Thus, the L1 norm is utilized to measure the consistency before and after image generation.

\subsubsection{Total Loss}
The total SIGAN loss includes adversarial loss, cycle consistency loss and SI loss, it is defined as follows:

\begin{equation}\label{eq3}
	\begin{split}
		\begin{array}{l}\begin{array}{lc}\begin{array}{l}L_{SIGAN}(G,\;F,\;D_a,\;D_b)\\=L_{adv}(G,\;D_b,\;A,\;B)+L_{adv}(F,\;D_a,\;B,\;A)\\+\;\lambda_1L_{SI}(G,\;A,\;B)+\lambda_1L_{SI}(F,\;A,\;B)\\+\;\lambda_2L_{cyc}(G,\;F)\\\end{array}&\\&\end{array}\\\end{array}
	\end{split}
\end{equation}
where terms $\lambda_1$ and $\lambda_2$ represent the relative importance between  adversarial loss, cycle consistency loss and SI loss. In the total loss, $L_{adv}$ guarantees that one domain can be transformed to another domain. $L_{cyc}$ ensures that the two domains can be converted to each other (defect2defect-free and defect-free2defect). $L_{SI}$ makes the generated image similar to the input image. Noted that domain $A$ and domain $B$ must have similar data distribution, which is an important precondition for the application of the proposed method.

\subsubsection{Generator Network}
\begin{figure}[!t]
	\centering
	\includegraphics[width=8cm]{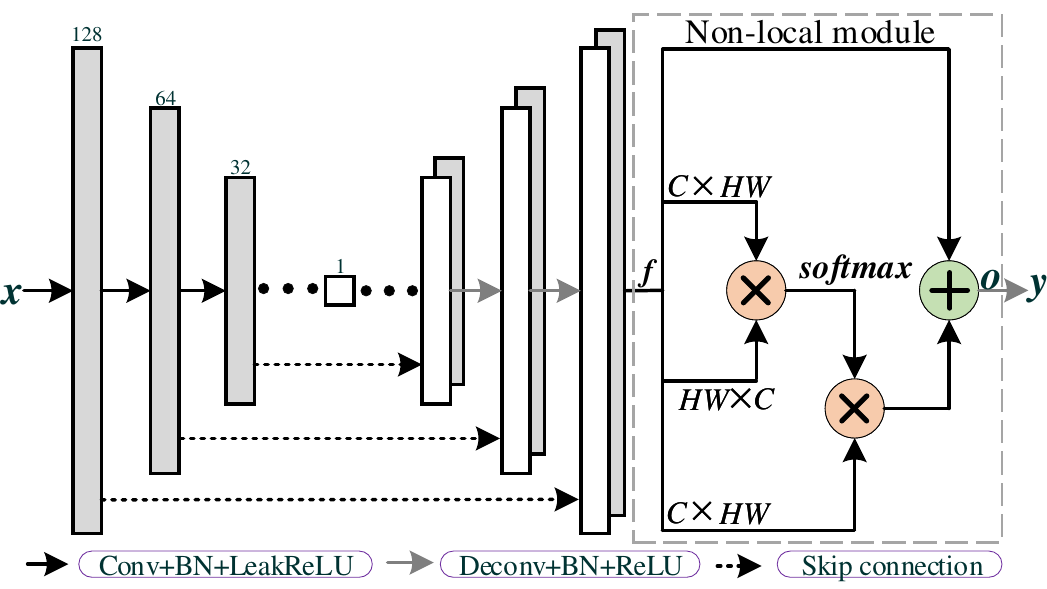}
	\caption{UNet-based architecture of generator $G$ and $F$, where $x$ is the input image, $y$ is the generated image.}\label{fig4}
\end{figure}

Many previous works \cite{Isola2017,Nguyen2020} adopts UNet \cite{Ronneberger2015} as the network architecture of the generator, which is also employed by $G$ and $F$ simultaneously. As shown in Fig. \ref{fig4}, UNet is an encoder-decoder network. The input image feature is extracted as the layer downsamples until a bottleneck, then the process is reversed. Moreover, the low-level layers include much more textural features, high-level layers include much more semantic features, skip connection can ensure that the high-level layers contain rich texture and semantic features simultaneously.   

Furthermore, we introduce the non-local module \cite{Wang2018} in the last of UNet to help the generator reconstructing realer image. The non-local module has achieved successful application to capture the global context information for pixel-based image generation \cite{Zhang2019, Mi2020, Wu2020}. Each pixel is generated by depending on the relationship with all other pixels. It is defined as: 

\begin{equation}\label{eq6}
	o=softmax(f\cdot f^T)\cdot f+f
\end{equation}
where $f$ is the input feature, $o$ is the output feature of the non-local module.

\subsubsection{Discriminator Network}
The discriminator network contains five convolution layers, which are employed to classify the generated image (real or fake). The purpose of discriminator is to avoid being cheated by generator. As shown in Fig. \ref{fig5}, discriminator $D_a$ expects to divide the generated image $F(b)$ with real image $a$. And discriminator $D_b$ expects to divide the generated image $G(a)$ with real image $b$. Table \ref{table1} presents the network details of the discriminator, which has five convolutional (conv) layers. Batch normalization (BN) and LeakyReLU are applied to normalize and activate the convolutional layers respectively. This discriminator has less computational complexity, and is effective to divide the image. Moreover, the network architectures of discriminators $D_a$ and $D_b$ share the same network, as mentioned above. 

\begin{table}[] 
	\renewcommand\arraystretch{1.2}
	\caption{Discriminator Network.}
	\centering
	\label{table1}
	\begin{tabular}{lcc}
		\hline
		Layer       & Filter size                                      & Stride \\ \hline
		Conv+BN+LeakyReLU & 64$\times$4$\times$4  & 2      \\
		Conv+BN+LeakyReLU & 128$\times$4$\times$4 & 2      \\
		Conv+BN+LeakyReLU & 256$\times$4$\times$4 & 2      \\
		Conv+BN+LeakyReLU & 512$\times$4$\times$4 & 1      \\
		Conv & 1$\times$4$\times$4   & 1      \\ \hline
	\end{tabular}
\end{table}

\subsection{Defect Segmentation Approach Based on defect2defect-free}
\begin{figure}[!t]
	\centering
	\includegraphics[width=8cm]{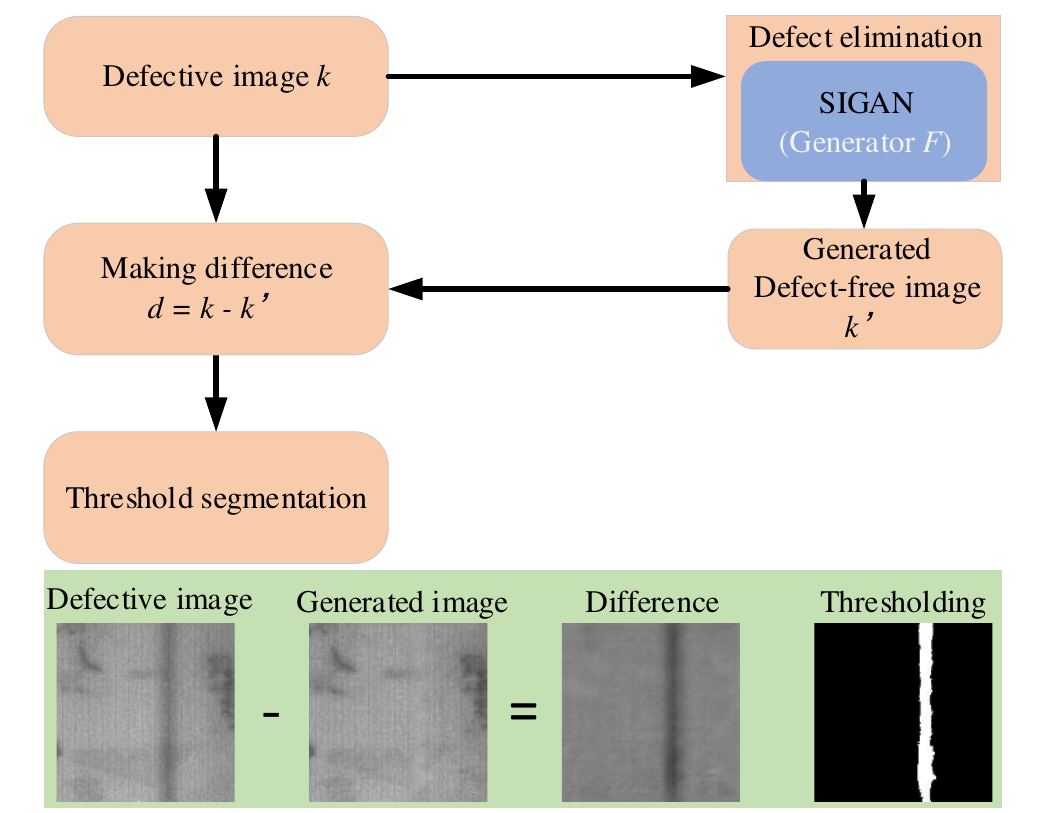}
	\caption{Pipeline of the proposed defect segmentation approach.}\label{fig6}
\end{figure}

As shown in Fig. \ref{fig6}, when inputting a defective image to generator $F$, a defect-free image is output. Generator $F$ is similar as a filter, which filters out the defect region, and retains the background region simultaneously. Then, the defect region can be acquired by employing the input defective image to subtract the output defect-free image. By setting an appropriate gray intensity threshold, the defect region can be detected. There is an important premise that the background area cannot be changed a lot during the image generation process. Then, the background can be eliminated by subtraction. To keep the background almost unchanged, the proposed SI loss plays a vital role to constrain it. The adversarial loss magnifies the difference between the generated image and the original input image to accomplish the domain transformation. However, the SI loss makes the difference smaller, and finally achieves a compromise balance with the adversarial loss. 

\subsection{Defect Augmentation Based on defect-free2defect}
\begin{figure}[!t]
	\centering
	\includegraphics[width=8cm]{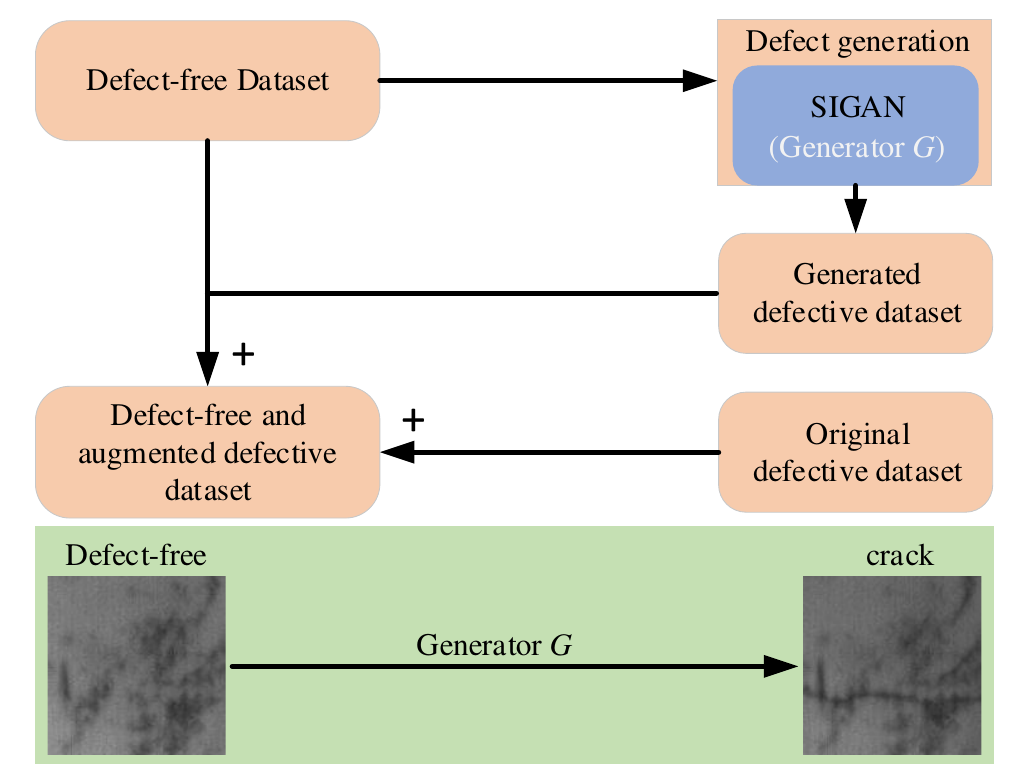}
	\caption{Pipeline of the defect augmentation approach.}\label{fig7}
\end{figure}

Generator $G$ can transform the defect-free image into defective image, which can be used to augment the small-samples dataset.  
As is known to all, deep CNN model is a data-driven approach \cite{Su2020}, which requires a great number of dataset for training. However, it is very difficult to obtain the defective images in the industrial manufacturing process. Thus, using defect-free images to generate defective images is of great research value. Our method provides a solution, which can employ the easily available defect-free images to generate defective images under complex background disturbance. As shown in Fig. \ref{fig7}, when inputting a defect-free image with random textural distribution, the generator $G$ will output an defective crack image. It is difficult to distinguish real or fake with the naked eye. To validate the effectiveness of the generated dataset, we utilize the generated defective dataset combining with original dataset to train and test the classification model. The detailed experimental results are presented in Section \uppercase\expandafter{\romannumeral4}. F. 

\section{Experiments}
In this section, several experiments are carried out to evaluate the performance of proposed SIGAN. Firstly, we introduce the dataset distribution. Secondly, the evaluation metrics are presented, which can quantify the experimental results. Thirdly, we introduce the detailed setting of the experiments. Finally, the experimental evaluations of image generation, defect segmentation, and defect augmentation are presented respectively.  

\subsection{Dataset}
\begin{figure}[!t]
	\centering
	\includegraphics[width=8cm]{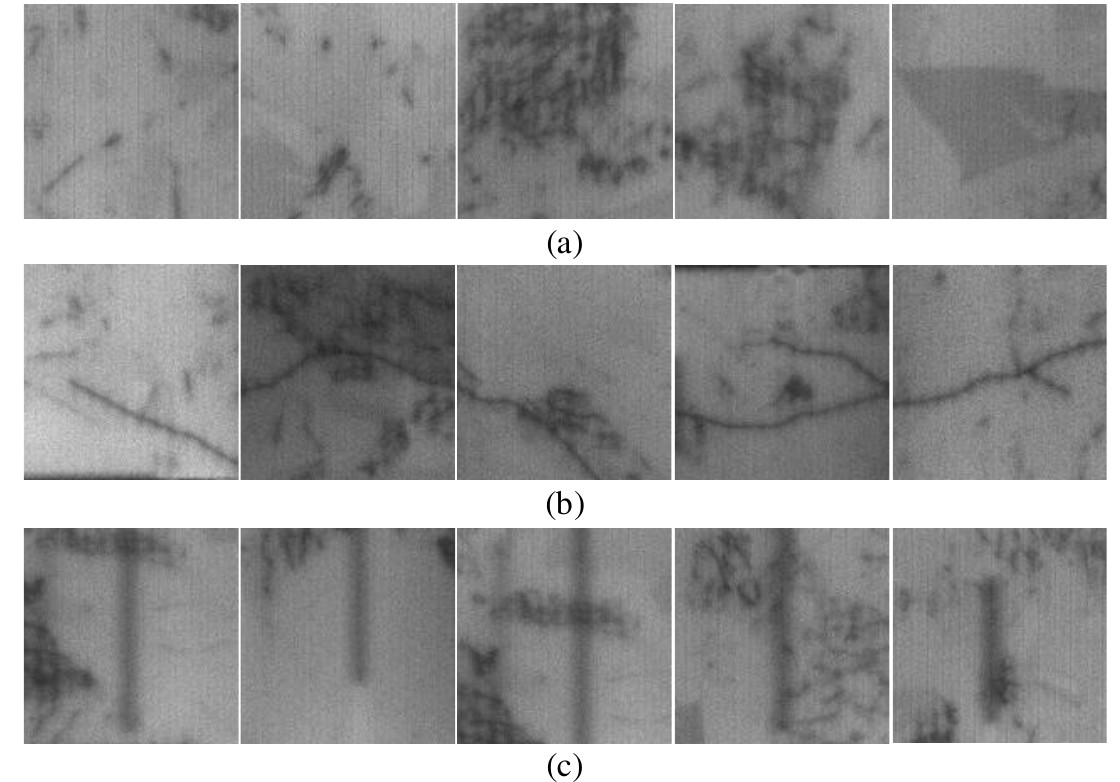}
	\caption{Various defect-free and defective images in EL dataset. (a) defect-free images. (b) crack defect. (c) finger interruption defects.}\label{fig8}
\end{figure}

The dataset used to evaluate the proposed method is collected in the manufacturing process of the multicrystalline solar cell. The solar cell EL image is collected by a near-infrared camera of WP-US146 with a SONY ICX825 chip. The raw image with a resolution of 1024$\times$1024 pixels is cropped into patches with a resolution of 128$\times$128 pixels. The categories of defects cover two types, i.e., crack and finger interruption. As shown in Fig. \ref{fig8}, crack defect appears as linear shape and randomly distributes in the image. Finger interruption defect presents cylindrical shape and is vertically located in the image. These two types of defects are easily disturbed by the complex background that shows dark and irregular area. Moreover, as shown in Fig. \ref{fig8}. (a), the defect-free image is also with the heterogeneous complex background disturbance.

The dataset distribution is illustrated in Table \ref{table2}. This dataset is employed to evaluate the performance of the image generation and defect segmentation tasks, we name it as $EL\_gen\_seg$  (EL-2019). For image preprocessing, the defective image dataset is insufficient to train a general model, and less training data will lead to over fitting. Thus, we use mirror, flip, and contrast normalization to augment the defective image dataset. Moreover, we access this solar cell EL image dataset at: \url{https://github.com/binyisu/EL-2019}. 

\begin{table}[]
	\renewcommand\arraystretch{1.3}
	\caption{$EL\_gen\_seg$ Distribution for the Evaluation of Image Generation and Segmentation.}
	\centering
	\label{table2}
	\begin{tabular}{lccc}
		\hline
		Dataset& Defect-free & Crack & Finger interruption \\ \hline
		Train & 200        & 50  & 50                \\ \cline{1-4}
		Test  & 80         & 80   & 80                 \\ \hline
	\end{tabular}
\end{table}

\subsection{Evaluation Metrics}
The Frechet Inception Distance score \cite{Heusel2017}, or FID for short, is a metric that calculates the distance between feature vectors calculated for real and generated images.
The score summarizes how similar the two groups of image features are. These features are calculated using the inceptionv3 \cite{Szegedy2016} model that is used for image classification. Lower scores indicate that the two groups of images are more similar, or have more similar statistics. A perfect score being 0.0 illustrates that the two groups of images are identical. The FID score is used to evaluate the quality of images generated by GAN, and lower scores have been shown to correlate well with higher quality images.

\begin{equation}\label{eq7}
	FID(x,\;g)=\left|\left|\mu_x-\mu_g\right|\right|_2^2+T(\phi_x+\phi_g-2(\phi_x\phi_g)^{1/2})
\end{equation}
where $T$ is the sum of all elements on the diagonal of the feature matrix. The mean value is $\mu$ and the covariance is $\phi$. Moreover, $x$ and $g$ represent the real image feature and the generated image feature extracted by pre-trained incptionv3 model.

\begin{table}[]
	\renewcommand\arraystretch{1.1}
	\caption{Evaluation Metrics of Image Segmentation.}
	\centering
	\label{table4}
	\begin{tabular}{ll}
		\hline
		Metrics   & Meaning                                                  \\ \hline
		$M_g$     & the number of defect pixels in ground truth              \\
		$M_d$     & the number of pixels detected by the segmentation method \\
		$M$       & the number of the same pixels in $M_g$ and $M_d$         \\
		$cpt$     & $cpt=M/M_g$                                              \\
		$crt$     & $crt=M/M_d$                                              \\
		$F-score$ & $F-score=(2\times cpt\times crt)/(cpt+crt)$              \\ \hline
	\end{tabular}
\end{table}

To evaluate the performance of defect segmentation, completeness (cpt), correctness (crt) and F-score are introduced. As illustrated in Table \ref{table4}, $M_g$ denotes the number of defect pixels in ground truth. $M_d$ represents the number of pixels detected by the segmentation algorithm. $M$ is the number of the same pixels in $M_g$ and $M_d$. Term $cpt$ illustrates the completeness of the segmentation method. Term $crt$ indicates the correctness of the segmentation results. The $F-score$ index can comprehensively evaluate the performance of segmentation algorithm, which
represents the weighted harmonic average of the $crt$ and $cpt$ indicators. The higher the value, the more effective the segmentation algorithm is.

\subsection{Implementation Details}
The proposed method is validated on a server with a Intel Core i7-10700 CPU and a NVIDIA GeForce RTX3090. The batch size is set to 4. In the total loss of SIGAN, the balanced terms $\lambda_1$ and $\lambda_2$ are set to 10 and 5 respectively. The learning rate 0.0002 is used to train all networks, and remains unchanged before the first 30 epochs, and then linearly decreased to zero after next 30 epochs.  The parameter optimization approach is adaptive moment estimation (Adam) algorithm \cite{Adam2014}, which can dynamically adjust the learning rate to prevent parameter oscillation. Moreover, before training, all images are normalized and resized to 256$\times$256 pixels.

\subsection{Image Generation}
Image generation based on the proposed SIGAN plays a vital role in the following two defect inspection tasks: defect segmentation based on defect2defect-free and defect augmentation based on defect-free2defect, as shown in Fig. 6 and Fig. 7. The following subsection quantifies the quality of image generation using FID indicator.   

\begin{figure}[!t]
	\centering
	\includegraphics[width=8.8cm]{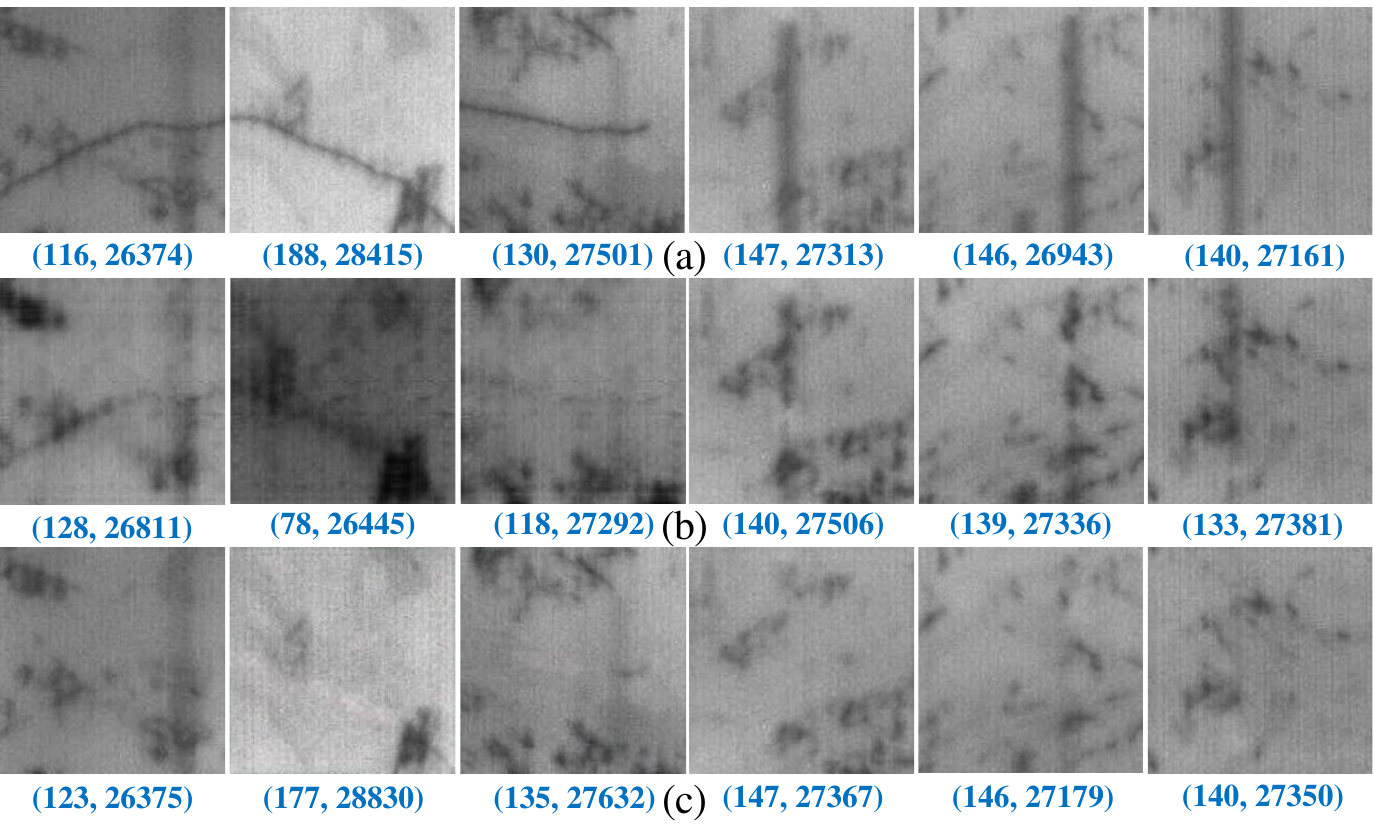}
	\caption{Defect2defect-free of SIGAN ($L_{adv}$+$L_{cyc}$) and SIGAN ($L_{adv}$+$L_{cyc}$+$L_{SI}$). (a) original defective images. (b) defect-free images generated by SIGAN ($L_{adv}$+$L_{cyc}$). (c) defect-free images generated by SIGAN ($L_{adv}$+$L_{cyc}$+$L_{SI}$). The blue values are the brightness and contrast of the image.}\label{fig9}
\end{figure}

\begin{figure}[!t]
	\centering
	\includegraphics[width=8.8cm]{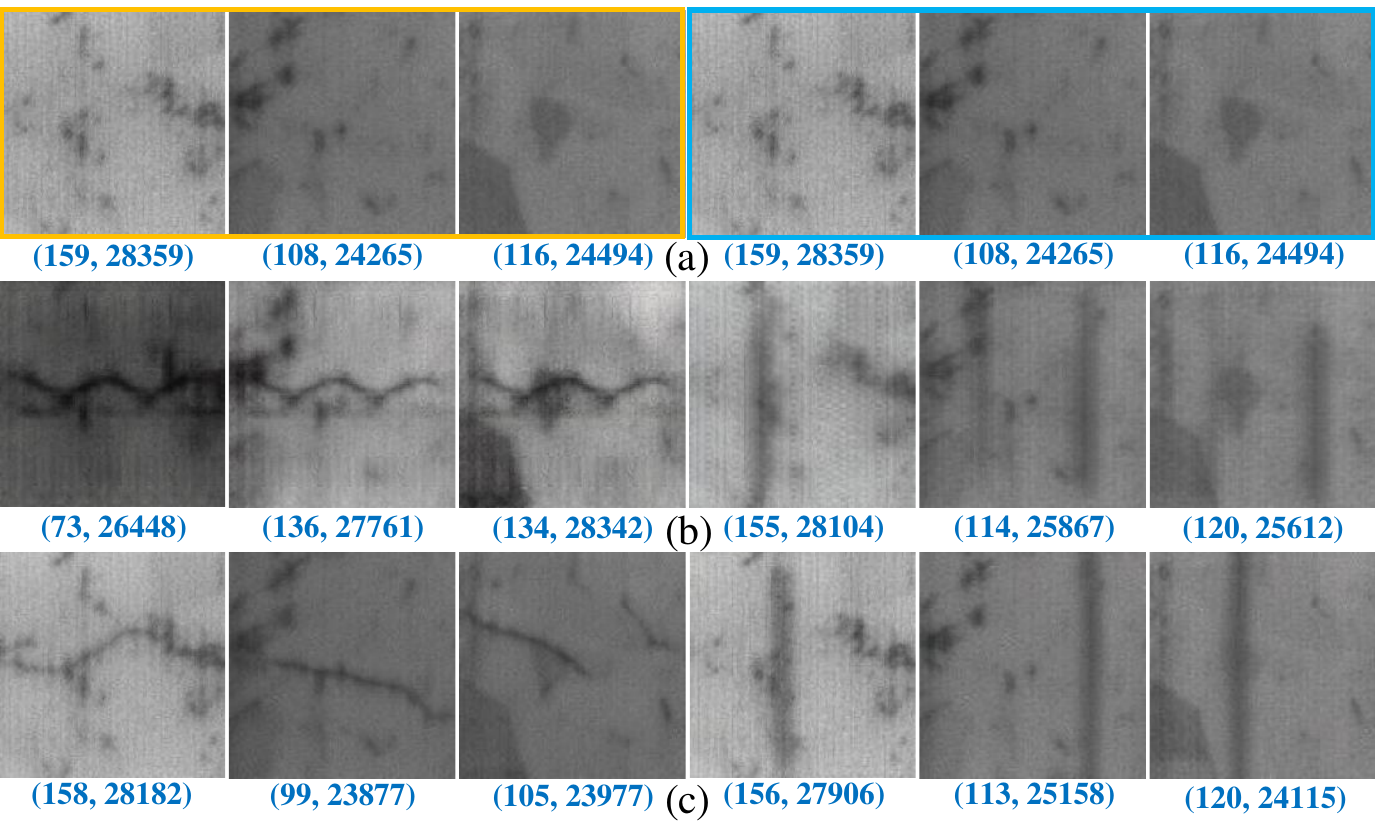}
	\caption{Defect-free2defect of SIGAN ($L_{adv}$+$L_{cyc}$) and  SIGAN ($L_{adv}$+$L_{cyc}$+$L_{SI}$). (a) original defect-free images. (b) defective image generation of SIGAN ($L_{adv}$+$L_{cyc}$).  (c) defective image generation of SIGAN ($L_{adv}$+$L_{cyc}$+$L_{SI}$). }\label{fig10}
\end{figure}

\subsubsection{Defect2defect-free}
The proposed SIGAN is compared with different loss functions for defect-free image generation on $EL\_gen\_seg$ dataset. As illustrated in Table \ref{table5}, by introducing the strong identity loss ($L_{SI}$) in  SIGAN ($L_{adv}$+$L_{cyc}$) (CycleGAN \cite{Zhu2017} is equivalent to SIGAN ($L_{adv}$+$L_{cyc}$).), the FID score of defect-free image generation is improved from 102.88 to 86.33. The reason is that the proposed SI loss makes the background of two image domains tend to be consistent, including color, texture, brightness, contrast and so on. The defective image background is retained, only the defect area is removed, which greatly improves the authenticity of the generated defect-free image. Thus, the SIGAN ($L_{adv}$+$L_{cyc}$+$L_{SI}$) achieves a better performance than SIGAN ($L_{adv}$+$L_{cyc}$). 

Fig. \ref{fig9} is used to qualitative analysis of the visualization improvement. Fig. \ref{fig9} (a), (b), and (c) are the original defective images, the defect-free images generated by SIGAN ($L_{adv}$+$L_{cyc}$), and the defect-free images generated by SIGAN ($L_{adv}$+$L_{cyc}$+$L_{SI}$) respectively. Comparing (a) with (b), it is not hard to see that when removing the defective region, the background is also changed. In the absence of SI loss constraint, SIGAN ($L_{adv}$+$L_{cyc}$) is easier to play freely, which leads to changes in the background of the generated image. However, comparing (a) with (c), the background texture is almost completely retained, meanwhile, the defect is almost completely removed. Furthermore, the brightness and contrast are closer before and after image generation, which proves the effectiveness of SI loss in the proposed SIGAN. Moreover, defect2defect-free is the basis for defect segmentation.

\begin{table}[]
	\renewcommand\arraystretch{1.2}
	\caption{FID Scores of Different Loss Functions for Image Generation. SIGAN ($L_{adv}$+$L_{cyc}$) is equivalent to the CycleGAN \cite{Zhu2017}.}
	\centering
	\label{table5}
	\begin{tabular}{lccc}
		\hline
		& Defect-free & Crack & Finger interruption \\ \hline
		SIGAN ($L_{adv}$+$L_{cyc}$)  & 102.88         & 156.25   & 91.12                 \\\cline{1-4} 
		SIGAN ($L_{adv}$+$L_{cyc}$+$L_{SI}$)             & \textbf{86.33}          & \textbf{100.05}    & \textbf{77.84}                  \\ \hline
	\end{tabular}
\end{table}

%
\begin{table}[]
	\renewcommand\arraystretch{1.4}
	\caption{Experimental Results of Five Methods.}
	\centering
	\label{table6}
	\begin{tabular}{llccc}
		\hline
		\multicolumn{1}{l}{Defect type}      & Method   & cpt   & crt   & F-score \\ \hline
		\multirow{5}{*}{Crack}               & FFT      & 64.54 & 44.47 & 52.67   \\ \cline{2-5} 
		& Gabor    & 60.49 & 50.30 & 54.92   \\ \cline{2-5} 
		& SEF      & 72.47 & 71.89 & 72.18   \\ \cline{2-5} 
		& CycleGAN & 55.62 & 10.66 & 17.89   \\ \cline{2-5} 
		& SIGAN   & \textbf{89.58} & \textbf{84.49} & \textbf{86.96}   \\ \hline
		\multirow{5}{*}{Finger interruption} & FFT      & 40.86 & 36.03 & 38.30   \\ \cline{2-5} 
		& Gabor    & 47.65 & 65.07 & 55.02   \\ \cline{2-5} 
		& SEF      & 66.92 & 65.04 & 65.97   \\ \cline{2-5} 
		& CycleGAN & 74.13 & 54.38 & 62.74   \\ \cline{2-5} 
		& SIGAN   & \textbf{94.47} & \textbf{92.96} & \textbf{93.71}   \\ \hline
		\multirow{5}{*}{Total}               & FFT      & 52.70 & 40.25 & 45.49   \\ \cline{2-5} 
		& Gabor    & 54.07 & 57.69 & 54.97   \\ \cline{2-5} 
		& SEF      & 69.69 & 68.47 & 69.08   \\ \cline{2-5} 
		& CycleGAN & 64.87 & 32.52 & 40.32   \\ \cline{2-5} 
		& SIGAN   & \textbf{92.02} & \textbf{88.73} & \textbf{90.34}   \\ \hline
		
	\end{tabular}
\end{table}

\begin{figure*}[!t]
	\centering
	\includegraphics[width=16cm]{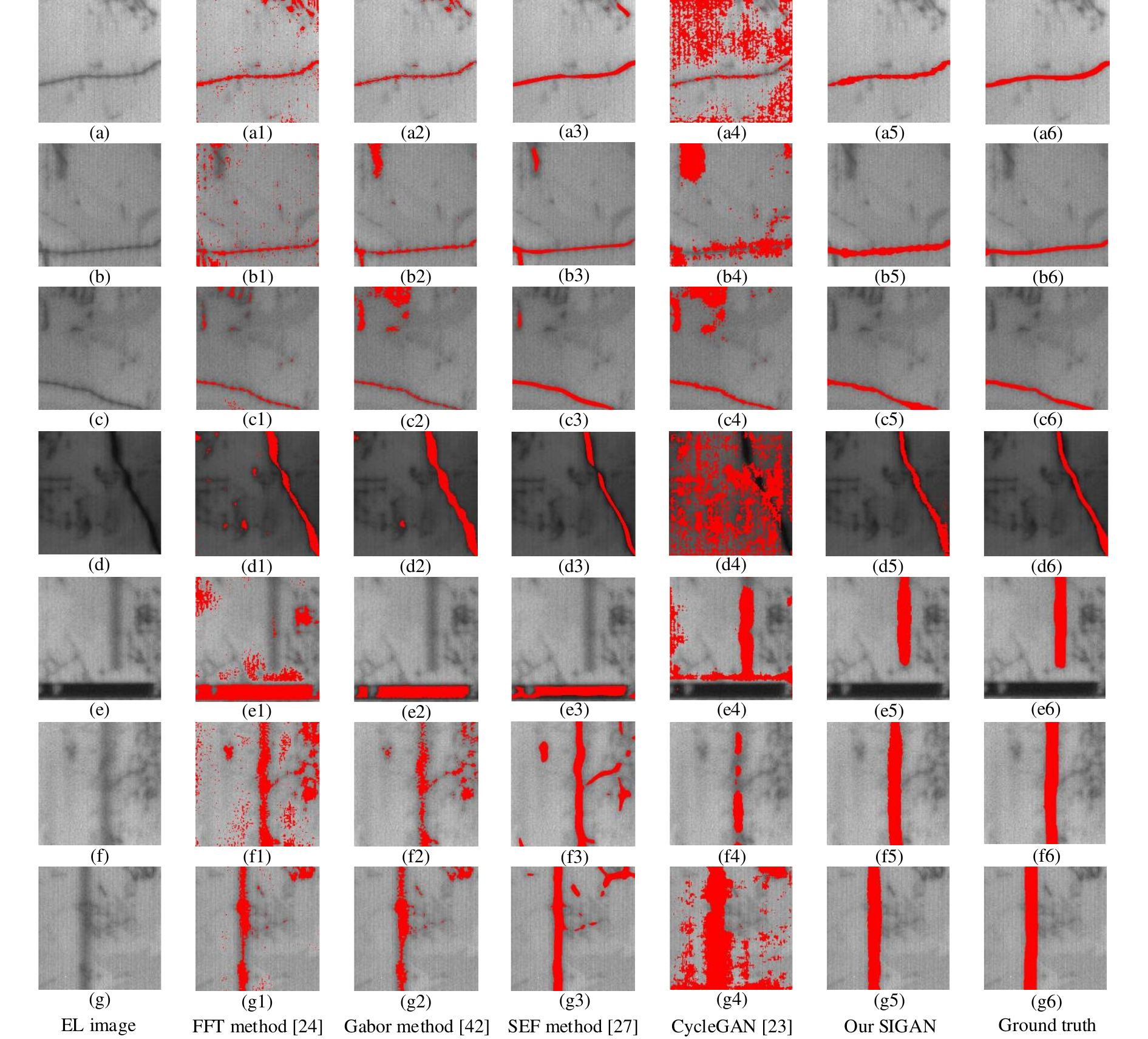}
	\caption{Detection results using the five methods. (a)-(g) EL images, (a1)-(g1), (a2)-(g2), (a3)-(g3), (a4)-(g4) and (a5)-(g5) corresponding detection results using FFT method, Gabor method, SEF method, CycleGAN, and our SIGAN. }\label{fig11}
\end{figure*}

\subsubsection{Defect-free2defect}
In terms of crack and finger interruption defective image generation, as illustrated in Table \ref{table5}, the FID scores of crack and finger interruption defective image generation are improved from 156.25, 91.12 to 100.05, 77.84 respectively. It indicates that SIGAN ($L_{adv}$+$L_{cyc}$+$L_{SI}$) generates realer image than SIGAN ($L_{adv}$+$L_{cyc}$). Moreover, as shown in Fig. \ref{fig10}, we employ the same three defect-free images (yellow box and blue box) to generate crack and finger interruption defects. For crack defect, the generated defective images of SIGAN ($L_{adv}$+$L_{cyc}$) are short of diversity. Otherwise, the brightness and contrast are changed a lot. However, the proposed SIGAN avoids these problems, which can add the defect region while keeping the image background almost unchanged. It shows that it is feasible to use a defect-free image to generate a relatively real defective image, which is of great significance for the augmentation of small-samples data in the industrial defect inspection process. 

\subsection{Defect Segmentation Based on Defect2defect-free}

In terms of defect segmentation, the proposed SIGAN can remove the defect region and almost completely retain the background region simultaneously. Then, the defect region is obtained by employing the input defective image to subtract the generated defect-free image. The defect detection process is similar to filter-based spectral domain methods, which can inspect the defect by the spectrum response. 

\subsubsection{Qualitative Evaluation}
The qualitative evaluation is shown in Fig. \ref{fig11}, the novel SIGAN is compared with Tsai's Fast Fourier Transform (FFT) method \cite{Tsai2012}, Gabor filter \cite{Gabor}, Chen's Steerable Evidence Filter (SEF) method \cite{Chen2020}, and CycleGAN \cite{Zhu2017}. To achieve a better performance of the four comparison methods, some key parameters are adjusted. Moreover, seven defective images with different shape, contrast and brightness are presented in the first column of Fig. \ref{fig11}, (a)-(d) are crack images, (e)-(g) are finger interruption images. The segmentation results are shown in the following five columns. Finally, the ground truths are listed in the last column.

Tsai's FFT method assumes that the defect texture is high frequency information, the background is low. However, some background textures are also high-frequency, such as the dark strip in the bottom of Fig. \ref{fig11} (e1) and the heterogeneous background in Fig. \ref{fig11} (f1). Thus, FFT method is easy to detect background as defect. The Gabor method is a linear filter for edge detection. The frequency and direction expression of Gabor filter is similar to that of human visual system. It is found that Gabor filter is very suitable for textural edge representation and separation. To achieve a better segmentation performance, we employ the convolution kernel with 8 directions and 6 scales to filter the image. Because the edge of crack defect is clear, the crack segmentation effect is better than FFT. However, as shown in Fig. \ref{fig11} (f2), the edge of finger interruption defect is frequently confused with the background, thus the background is easy to be segmented as defect. The segmentation results of Gabor are still not satisfactory. In terms of filter-based methods, Chen's SEF achieves the best segmentation effect. SEF is an efficient oriented filter. To improve the robustness of intensity variation, except for the direction, the spatial distance is also taken into account. As shown in Fig. \ref{fig11} (d2), in the case of weaker intensity change, SEF method performs better than other filters. Moreover, as shown in Fig. \ref{fig11} (f3) and (g3), the segmentation performance is better than FFT and Gabor for the finger interruption defects with weak edge intensity change. However, some backgrounds are still easy to be detected as defects.

The CycleGAN method also uses the difference between the images before and after generation to achieve defect segmentation. Due to the absence of identity loss constraint, the background is changed too much, the results are shown in Fig. \ref{fig11} (a4)-(g4). Compared with above four approaches, the novel SIGAN can extract the defect region more accurate and complete without the background disturbance. The experimental results are presented in Fig. \ref{fig11} (a5)-(g5). Compared with FFT, Gabor, SEF and CycleGAN, SIGAN can better remove the complex background region, and retain more complete defect area, which verifies its effectiveness for defect segmentation.

\subsubsection{Quantitative Evaluation}
The quantitative evaluation is presented in Table \ref{table6}. This experiment evaluates two types of defects (crack and finger interruption) with five approaches. The evaluation metrics (cpt, crt, and F-score) are given to assess the performance of different methods. For Tsai's FFT method, it performs badly for the finger interruption defect inspection (38.30\% F-score), this is because the gray intensity of finger interruption is close to the complex background, thus it is very easy to confuse the FFT filter. To achieve a good result, Gabor filter uses a convolution kernel of 8 directions and 6 scales to filter the image, this filter has a well response to the edge gradient. However, the edge gradient of complex background is similar as the defect gradient, thus it also achieves a poor segmentation result (54.97\% total F-score). Steerable evidence filter (SEF) \cite{Chen2020} is proposed in recent years, it has a 69.08\% total F-score that is higher than other filters. The linear crack defect can be highlighted in the surrounding complex background. SEF responds strongly to linear textures, thus the experimental performance is better. 

\begin{figure}[!t]
	\centering
	\includegraphics[width=7.8cm]{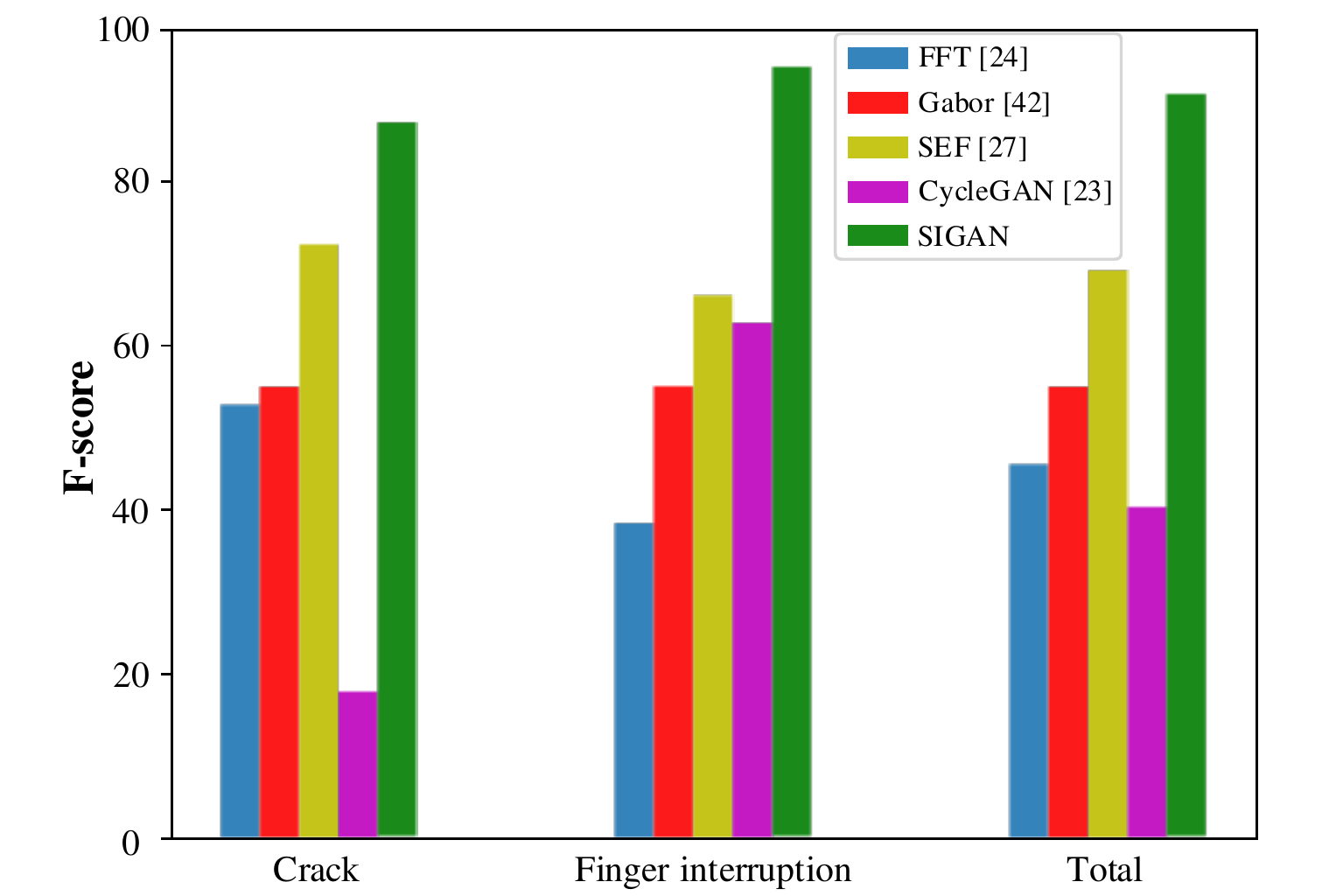}
	\caption{F-score values of different methods.}\label{fig12}
\end{figure}

For CycleGAN, the segmentation result of crack defect is poor (17.89\% F-score), the reason is that as shown in Fig. \ref{fig9}, although the crack defect can be partially removed, the background texture of the generated defect-free image is changed a lot, thus the defect segmentation results contain much more background region, which will reduce the defect inspection effect. As for the proposed SIGAN, the strong identity loss will ensure that the background before and after generation is as similar as possible, thus SIGAN can achieve a good performance to keep the background almost unchanged while removing the defects. The total cpt, crt and F-score of SIGAN are 92.03\%, 88.73\% and 90.34\% respectively, which is better than other four methods. The above experimental results validate the effectiveness of the proposed method. Moreover, the F-score values of different methods are shown in Fig. \ref{fig12}.

\subsubsection{Time Efficiency}
The time efficiency evaluation is presented in Table \ref{table8}. Time efficiency evaluation is conducted on a server with a Intel Core i7-10700 CPU and a NVIDIA GeForce RTX3090. Noting that GAN-based approaches call the GPU to process the image. Owing to the acceleration of GPU, the speed of SIGAN is 62 ms that is not slower than other methods.  

\begin{table}[]
	\renewcommand\arraystretch{1.3}
	\caption{Time Efficiency Evaluation.}
	\centering
	\label{table8}
	\begin{tabular}{lccccc}
		\hline
		Method            & FFT & Gabor & SEF & CycleGAN & SIGAN \\ \hline
		Average time (ms) & 72  & 94    & 86  & 62       & \textbf{62}    \\ \hline
	\end{tabular}
\end{table}



\begin{table}[]
	\renewcommand\arraystretch{1.2}
	\caption{$EL\_aug$ Dataset Distribution.}
	\centering
	\label{table3}
	\begin{tabular}{l|c|cc|cc}
		\hline
		\multicolumn{1}{c|}{\multirow{2}{*}{Generated dataset}} & \multicolumn{1}{c|}{\multirow{2}{*}{Defect-free}} & \multicolumn{2}{c|}{Crack} & \multicolumn{2}{c}{Finger interruption} \\ \cline{3-6} 
		\multicolumn{1}{c|}{}                                     & \multicolumn{1}{c|}{}                             & real   & fake              & real         & fake                     \\ \hline
		Train                                                     & 200                                              & 50   & 200              & 50         & 200                     \\ \hline
		Test                                                      & 80                                               & 80    & \textbackslash{}  & 80          & \textbackslash{}         \\ \hline
	\end{tabular}
\end{table}

\begin{table*}[]
	\renewcommand\arraystretch{1.3}
	\caption{Image Classification Results with different CNN models.}
	\centering
	\label{table7}
	\begin{tabular}{c|c|ccc|ccc}
		\hline
		    \multirow{2}{*}{Defect Type}     & \multirow{2}{*}{Method} & \multicolumn{3}{c|}{Original dataset} & \multicolumn{3}{c}{Augmentated dataset} \\ \cline{3-8}
		                                     &                         & Precision & Recall &    F-measure     & Precision & Recall &     F-measure      \\ \hline
		       \multirow{4}{*}{Crack}        &        Resnet50         &   96.33   & 96.35  &      96.34       &   97.58   & 97.91  &       97.74        \\ \cline{2-8}
		                                     &        Mobilenet        &   96.26   & 98.44  &      97.34       &   97.93   & 98.96  &       98.44        \\ \cline{2-8}
		                                     &       Inceptionv3       &   97.95   & 98.83  &      98.39       &   98.11   & 98.87  &       98.98        \\ \cline{2-8}
		                                     &       Densenet121       &   98.79   & 99.07  &      98.93       &   99.71   & 99.91  &       99.81        \\ \hline
		\multirow{4}{*}{Finger interruption} &        Resnet50         &   96.27   & 98.96  &      97.60       &   96.77   & 98.98  &       97.86        \\ \cline{2-8}
		                                     &        Mobilenet        &   96.13   & 99.36  &      97.72       &   97.79   & 99.08  &       98.43        \\ \cline{2-8}
		                                     &       Inceptionv3       &   97.86   & 99.48  &      98.66       &   98.70   & 99.31  &       99.00        \\ \cline{2-8}
		                                     &       Densenet121       &   98.72   & 99.67  &      99.19       &   99.79   & 100.00 &       99.89        \\ \hline
		       \multirow{4}{*}{Total}        &        Resnet50         &   96.30   & 97.66  &      96.97       &   97.18   & 98.45  &   \textbf{97.80}   \\ \cline{2-8}
		                                     &        Mobilenet        &   96.20   & 98.90  &      97.53       &   97.86   & 99.02  &   \textbf{98.44}   \\ \cline{2-8}
		                                     &       Inceptionv3       &   97.91   & 99.16  &      98.53       &   98.41   & 99.09  &   \textbf{98.99}   \\ \cline{2-8}
		                                     &       Densenet121       &   98.76   & 99.37  &      99.06       &   99.75   & 99.96  &   \textbf{99.85}   \\ \hline
	\end{tabular}
\end{table*}

\subsection{Defect Augmentation Based on Defect-free2defect}
As listed in Table \ref{table3}, the augmented dataset named as $EL\_aug$ is used to validate the defective image augmentation effect of SIGAN. The training data includes 50 real crack and 50 real finger interruption defects. We apply 200 defect-free images to generate 200 fake crack images and 200 fake finger interruption images, which are employed to augment the small-samples dataset. To validate the effectiveness of defective image augmentation with SIGAN, the augmented dataset is compared with the original dataset in terms of image classification task. The classification performance is evaluated by the indicators: precision, recall and F-measure \cite{Su2021}, which have similar means as the indicators cpt, crt and F-score respectively in Table \ref{table4}, except that pixels are replaced by images.

The experimental results are illustrated in Table \ref{table7}. Four CNN-based methods (Resnet50 \cite{Resnet2016}, Mobilenet \cite{Mobilenet}, Inceptionv3 \cite{Szegedy2016}, and Densenet121 \cite{Densenet}) are used to evaluate the image classification results. For different methods, the augmented dataset has a 0.83\%, 0.91\%, 0.46\% and 0.79\% higher total F-measure respectively. In the experiments, the method based on SIGAN for data augmentation significantly increases the recognition rate of the solar cell EL defects, mainly due to the following improvements from the data distribution and training process. Firstly, regarding the data distribution, with the assistance of data augmentation by the SIGAN, the border between the defect and defect-free data distributions becomes clearer, which can improve the recognition of data in the border area, such as data with poor defect features because of poor or uneven lighting. Images generated using the SIGAN possess better quality and diversity; Therefore, our proposed method makes the border clearer and further decreases the inaccuracy. Secondly, regarding the training process, the data
distribution after data augmentation based on SIGAN becomes more diverse. Therefore, the models do not easily over fitting during the training process, which increases the accuracy of recognition.

\section{Conclusion}
In this paper, a novel SIGAN is proposed to solar cell EL defect segmentation and augmentation without paired labeled image for training. We provide a new idea that applies GAN for industrial defect segmentation. SIGAN can accomplish defective image and defect-free image generation, and keep the complex background almost unchanged. The generated defect-free image can be made difference with the input defective image to realize defect segmentation. Moreover, the generated defective image can be used for solar cell EL image dataset augmentation, which can be employed to fine-tuning the CNN models. Experimental results show that the proposed SIGAN performs better than other methods in terms of defects segmentation in EL image, and is effective to augment the small-samples solar cell defect dataset simultaneously. 

The limitation of SIGAN is that this method can only generate EL image patch with 256$\times$256 pixels, how to generate raw EL image with 1024$\times$1024 pixels is our future research work.


%

%
%
%
%
%

\ifCLASSOPTIONcaptionsoff
  \newpage
\fi



%
\normalem

%

%
%
%
\vspace{-1.3cm}
\begin{IEEEbiography}[{\includegraphics[width=1in,height=1.25in,clip,keepaspectratio]{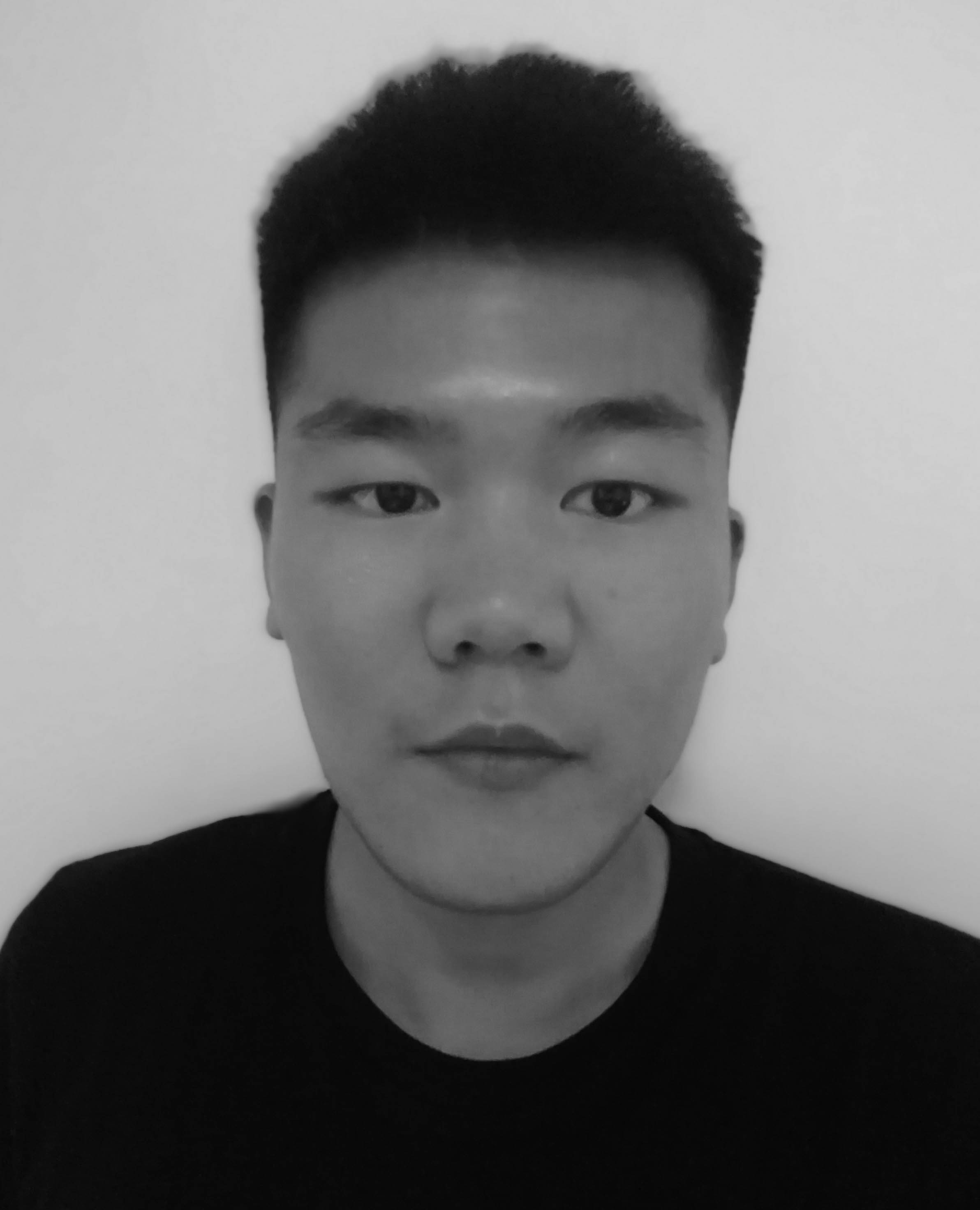}}]
	{Binyi Su} received the B.S. degree in intelligent 
	science and technology from the Hebei
	University of Technology, Tianjin, China, in 2017, and the M.S degree in control engineering from the Hebei
	University of Technology, Tianjin, China, in 2020.
	
	He is currently pursuing the Ph.D. degree in computer science and technology from Beihang University, Beijing, China.
	His current research interests include computer vision and pattern recognition, machine learning and artificial intelligence, industrial image defect detection.
\end{IEEEbiography}

\vspace{-1.3cm}
\begin{IEEEbiography}[{\includegraphics[width=1in,height=1.25in,clip,keepaspectratio]{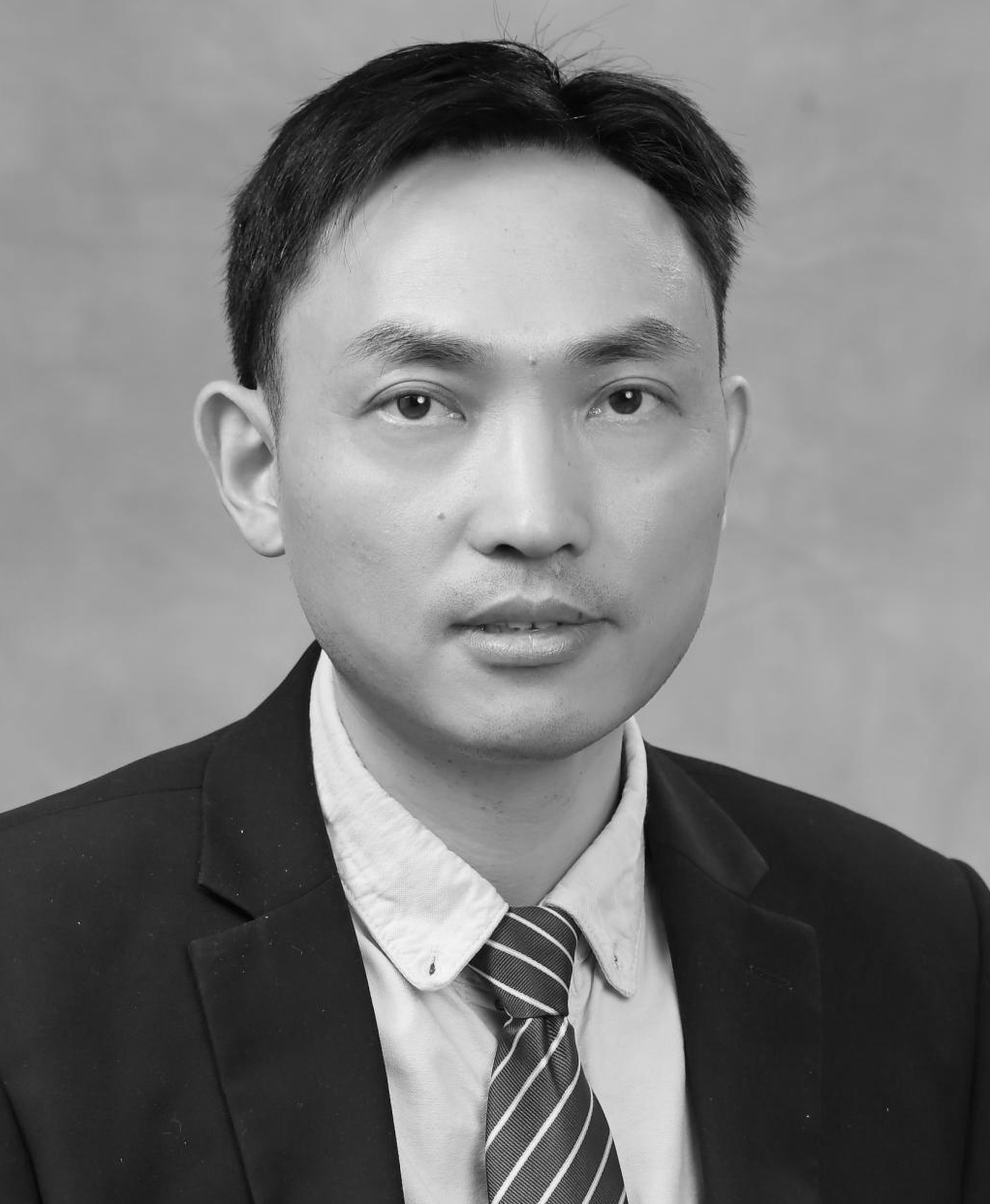}}]{Zhong Zhou} received the B.S. degree in material physics from Nanjing University in 1999 and the Ph.D. degree in computer science and technology from Beihang University, Beijing, China, in 2005. 
	
	He is currently a Professor and the Ph.D. Adviser with the State Key Laboratory of Virtual Reality Technology and Systems, Beihang University. His main research interests include virtual reality, augmented reality, computer vision, and artificial intelligence.
\end{IEEEbiography}

\vspace{-1.3cm}
\begin{IEEEbiography}[{\includegraphics[width=1in,height=1.25in,clip,keepaspectratio]{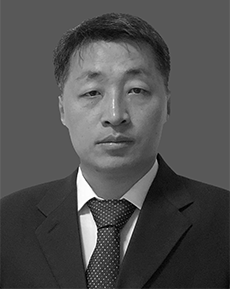}}]{Haiyong Chen}
	received the M.S. degree in detection technology and automation from the Harbin University of Science and Technology, Harbin, China, in 2005, and the Ph.D. degree in control science and engineering from the Institute of Automation, Chinese Academy of Sciences, Beijing, China, in 2008. 	
	
	He is currently a Professor with the School of Artificial Intelligence and Data Science, Hebei University of Technology, Tianjin. He is also an expert in the field of photovoltaic cell image processing and automated production equipment. His current research interests include image processing, robot vision, and pattern recognition.
\end{IEEEbiography}

\vspace{-1.3cm}
\begin{IEEEbiography}[{\includegraphics[width=1in,height=1.25in,clip,keepaspectratio]{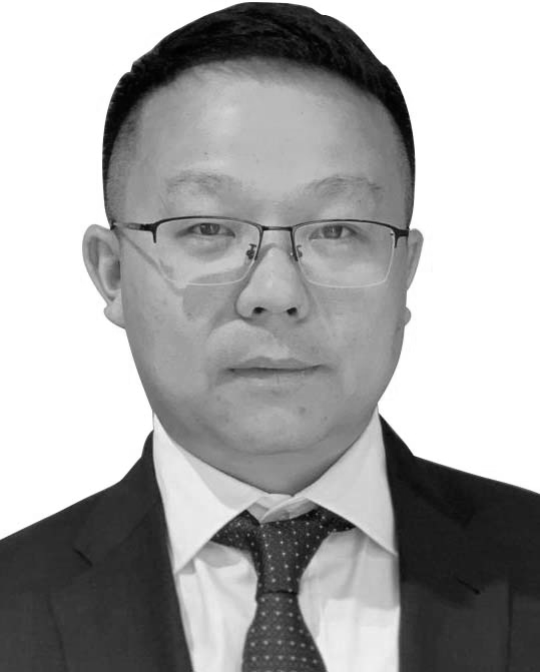}}]{Xiaochun Cao}(Senior Member, IEEE) received the B.E. and M.E. degrees in computer science from Beihang University, Beijing, China, and the Ph.D. degree in computer science from the University of Central Florida, Orlando, FL, USA. He has spent about three years with ObjectVideo Inc., as a Research Scientist. From 2008 to 2012, he was a Professor with Tianjin University, Tianjin, China. He has been a Professor with the Institute of Information Engineering, Chinese Academy of Sciences, Beijing, since 2012. He has authored or coauthored over 120 journal articles and conference papers. He is a fellow of the IET. His dissertation was nominated for the University of Central Florida’s University-level Outstanding Dissertation Award. In 2004 and 2010, he was a recipient of the Piero Zamperoni Best Student Article Award at the International Conference on Pattern Recognition. He is on the Editorial Board of the IEEE TRANSACTIONS ON IMAGE PROCESSING.
\end{IEEEbiography}



\end{document}